\documentclass[pdflatex,sn-mathphys-num]{sn-jnl}

\usepackage{graphicx}
\usepackage{multirow}
\usepackage{amsmath,amssymb,amsfonts}
\usepackage{amsthm}
\usepackage{mathrsfs}
\usepackage[title]{appendix}
\usepackage{xcolor}
\usepackage{textcomp}
\usepackage{manyfoot}
\usepackage{hyperref}
\usepackage{tikz}
\usetikzlibrary{arrows.meta,positioning,shapes.geometric,calc,fit}
\usepackage{pgfplots}
\pgfplotsset{compat=1.18}
\graphicspath{{figures/}}

\begin{document}

\title[Decoding Student Minds]{Decoding Student Minds: Leveraging Conversational Agents for Psychological and Learning Analysis}

\author*[1]{\fnm{Nour El Houda} \sur{Ben Chaabene}}\email{nour-el-houda.ben\_chaabene@sorbonne-universite.fr}

\author[2]{\fnm{Hamza} \sur{Hammami}}\email{hamza.hammami@enit.utm.tn}

\author[3]{\fnm{Laid} \sur{Kahloul}}\email{l.kahloul@univ-biskra.dz}

\affil*[1]{\orgdiv{STIH Laboratory}, \orgname{Sorbonne University}, \orgaddress{\street{28 Rue Serpente}, \city{Paris}, \postcode{75006}, \country{France}}}

\affil[2]{\orgdiv{LIPAH-LR11ES14}, \orgname{National Engineering School of Tunis, and Faculty of Sciences of Tunis}, \orgaddress{\street{Campus Universitaire Tunis El-Manar}, \city{Tunis}, \postcode{2092}, \country{Tunisia}}}

\affil[3]{\orgdiv{LINFI Laboratory}, \orgname{Biskra University}, \orgaddress{\city{Biskra}, \postcode{07000}, \country{Algeria}}}

\abstract{This paper presents a psychologically-aware conversational agent designed to enhance both learning performance and emotional well-being in educational settings. The system combines Large Language Models (LLMs), a knowledge graph-enhanced BERT (KG-BERT), and a bidirectional Long Short-Term Memory (LSTM) network with attention to classify students' cognitive and affective states in real time. Unlike prior chatbots limited to either tutoring or affective support, our approach leverages multimodal data-including textual semantics, prosodic speech features, and temporal behavioral trends-to infer engagement, stress, and conceptual understanding. A pilot study with 45 university students demonstrated improved motivation, reduced stress, and moderate academic gains compared to unimodal baselines. We explicitly discuss the exploratory nature of this small-sample pilot, report effect sizes and inter-rater reliability alongside significance tests, and provide an ablation analysis isolating the contribution of the knowledge-graph component and of each modality. These results underline the promise-while acknowledging the current limits in scale and modality balance-of integrating semantic reasoning, multimodal fusion, and temporal modeling to support adaptive, student-centered educational interventions.}

\keywords{Conversational Agents, Psychological Analysis, Student Behavior, Learning Process, LLM and KG-BERT}

\maketitle

\section{Introduction}\label{sec1}

In today's educational environments, academic performance and emotional well-being are increasingly seen as interdependent dimensions of student success. Factors such as stress, anxiety, and fluctuating motivation are now widely acknowledged as critical determinants of learning outcomes~\cite{dmello2012,camachoMorles2021}. Recent studies emphasize the importance of capturing multimodal and socio-emotional indicators to explain learning variability~\cite{llm_education_survey_2025}, highlighting the need for systems that can respond to students' psychological states in real time. Yet, most existing technologies treat cognition and emotion separately, leading to fragmented and often ineffective interventions.

The emergence of LLMs in educational applications opens new avenues for developing more holistic, context-aware support systems~\cite{nye2014}. When endowed with empathic capabilities, such systems have the potential to deliver emotionally intelligent and personalized feedback, fostering student motivation and mitigating stress~\cite{ortegaOchoa2024,liu2024,empathic_platform_jmir_2024}. Recent educational chatbot frameworks illustrate complementary strategies for improving intent recognition and response quality: Maheswari et al.~\cite{maheswari2023} propose an LSTM-based feature-extraction pipeline for teaching-learning support, while Anitha Elavarasi and Jayanthi~\cite{educhat2024} introduce a hybrid heuristic-optimized deep learning framework (E1DCNN-LSTM with XLNet/BERT fused features) that improves intent detection accuracy over BiLSTM, 1DCNN, and ensemble baselines. However, many current agents-including these-are limited to surface-level sentiment analysis or rely on rigid, rule-based logic, preventing them from modeling the complex and dynamic interplay between learners' affective and cognitive states.

This paper investigates the following research question: \textit{Can a psychologically-aware conversational agent-leveraging semantic reasoning and prosodic speech features-enhance both emotional well-being and academic engagement of students in real time, as evidenced by reduced stress and increased motivation?}

To address this, we present a multimodal architecture that combines semantic understanding and vocal emotion detection within a unified adaptive framework. The agent integrates (i) a domain-specific knowledge graph-enhanced BERT (KG-BERT) for contextualized semantic reasoning, and (ii) a prosodic analysis module that extracts vocal indicators such as pitch, intensity, and rhythm to assess emotional states. These signals are fused through a bidirectional Long Short-Term Memory network with attention mechanisms to jointly infer students' levels of engagement, stress, motivation, and conceptual understanding. This design is inspired by recent advances in hybrid LLM-KG systems~\cite{knowledge_graph_llm_embedding_survey_2024}, which combine symbolic and sub-symbolic reasoning for richer contextual modeling. Because interactions in the present pilot were predominantly textual (70\%) with optional spoken input (30\%), we report a dedicated modality ablation (Section~\ref{sec:ablation}) to clarify the actual contribution of the prosodic and knowledge-graph components rather than presenting the fusion architecture as an unconditionally validated contribution.

Unlike previous approaches focused exclusively on either academic scaffolding (e.g., AutoTutor~\cite{nye2014}) or emotional support (e.g., empathic conversational agents~\cite{ortegaOchoa2024}), our model offers a real-time, dual-focus monitoring mechanism grounded in multimodal fusion, evaluated here through an exploratory pilot study rather than a large-scale validation.

The main contributions of this work are as follows:
\begin{enumerate}
\item We design and implement a psychologically-aware conversational agent that combines LLM-based semantic reasoning, knowledge graph contextualization, and prosodic analysis for dynamic student state monitoring.
\item We conduct an 8-week pilot study involving 45 university students, demonstrating that the system is associated with reduced stress and anxiety, increased motivation, and improved academic engagement, and we report effect sizes and inter-rater reliability to support the interpretability of these findings.
\item We provide an ablation study isolating the contribution of KG-BERT augmentation and of each input modality (text vs. prosody) to the reported performance gains.
\item We reflect on the broader implications of this architecture for the development of next-generation educational systems, particularly in terms of personalization, ethical deployment, scalability, and inclusive design.
\end{enumerate}

The remainder of this paper is structured as follows: Section~\ref{RW} reviews related work on conversational agents, affective computing, and knowledge-graph-enhanced LLMs. Section~\ref{model} describes the system architecture and methodology. Section~\ref{RstDisc} presents the experimental results, ablation study, and discussion. Section~\ref{conc} concludes with key insights and directions for future research.

\section{Related Work}\label{RW}

Research on intelligent educational agents has evolved along five key axes: conversational tutoring systems, empathic and emotion-aware agents, semantic reasoning via knowledge graphs, multimodal affect recognition, and hybrid deep-learning chatbot frameworks. While each line of inquiry has yielded significant advances, relatively few efforts have successfully integrated these dimensions into a unified, psychologically-aware architecture.

\subsection{Conversational Tutoring Systems}
Early dialogue-based systems primarily targeted cognitive improvement. Fryer and Carpenter~\cite{fryer2006} showed that conversational agents could enhance student motivation, while Kerly et al.~\cite{kerly2007} reported improved engagement through emotionally framed prompts. However, these systems relied heavily on rule-based scripting and lacked adaptive capabilities or real-time psychological modeling.

\subsection{Hybrid Deep-Learning Educational Chatbots}
More recent educational chatbot research has focused on improving intent recognition and response accuracy through hybrid deep-learning pipelines. Maheswari et al.~\cite{maheswari2023} designed an educational chatbot that combines feature extraction with an LSTM-based intent classifier, reporting gains in teaching-learning effectiveness over conventional retrieval-based systems. Anitha Elavarasi and Jayanthi~\cite{educhat2024} proposed EduChatbot, which fuses XLNet and BERT features and optimizes an Enhanced 1D-CNN-LSTM (E1DCNN-LSTM) classifier with a bio-inspired hybrid heuristic (PA-BEPOCPA), achieving higher accuracy than BiLSTM, 1DCNN, LSTM, and ensemble baselines on their dataset. Complementary hybrid intent-recognition work combining fine-tuned BERT-family encoders with bidirectional LSTM networks and a confidence-based fallback mechanism has similarly improved intent-classification accuracy and reliability in service-oriented chatbot deployments~\cite{satya2026hybrid}. These contributions demonstrate the value of hybrid feature-fusion and optimization strategies for chatbot accuracy, but, unlike our approach, they do not address the joint modeling of learners' psychological (affective) states alongside cognitive intent, nor do they incorporate structured knowledge-graph reasoning or multimodal (voice) input.

\subsection{Emotion-Aware and Empathic Agents}
Recent developments have shifted toward incorporating affective support. Ortega-Ochoa et al.~\cite{ortegaOchoa2024} reviewed agents that aim to reduce stress and foster emotional regulation. Liu et al.~\cite{liu2024} demonstrated that combining knowledge scaffolding (KS) with emotional scaffolding (ES) improves both learning and well-being. Similarly, empathic platforms capable of delivering adaptive socio-emotional feedback have been shown to increase student motivation and persistence~\cite{empathic_platform_jmir_2024}. More recently, Kar et al.~\cite{mathbuddy2025} introduced MathBuddy, an emotionally aware LLM-powered math tutor that aggregates text- and facial-expression-based emotion signals and reports substantial gains in pedagogical response quality, while Prasad and Kanna~\cite{prasad2025} proposed a next-generation emotion-aware intelligent tutoring system combining facial, speech, and physiological signals with attention-based fusion. Nonetheless, these systems either omit semantic/knowledge-graph reasoning or, when multimodal, do not isolate each modality's contribution through ablation-an issue we directly address in Section~\ref{sec:ablation}.

\subsection{Semantic Reasoning and Knowledge Graph Integration}
To address limitations in factual grounding, several approaches now combine language models with structured knowledge. KG-BERT~\cite{yao2019} enhances BERT by incorporating knowledge graph embeddings, enabling deeper conceptual reasoning. Such hybrid models improve explainability and reduce hallucinations~\cite{knowledge_enhanced_agents_2024}, and recent work in adjacent domains (e.g., clinical diagnosis) confirms via explicit ablation that knowledge-injection modules meaningfully improve downstream accuracy over language-model-only baselines~\cite{kg_ablation_2025}. Yet the application of KG-BERT-style reasoning in learning contexts-particularly in combination with emotion modeling and with rigorous ablation of the knowledge component itself-remains limited~\cite{llm_education_survey_2025}, a gap this paper addresses through a dedicated KG-BERT ablation (Section~\ref{sec:ablation}). Related knowledge-graph-augmented tutoring frameworks provide converging evidence for the value of structured knowledge integration: KG-RAG~\cite{kgrag2023} reports a controlled 35\% gain in assessment scores ($p<0.001$, $n=76$) attributable to graph-augmented retrieval, and a recent multimodal-knowledge-graph tutoring system~\cite{kgrag_its_2026} similarly demonstrates automatic graph construction improving tutoring accuracy and responsiveness-both consistent with, and independently corroborating, our own KG-BERT ablation result (Table~\ref{tab:ablation}).

\subsection{Multimodal Emotion Recognition in Education}
Multimodal affective computing combines signals such as speech, text, and interaction patterns to infer emotional states. While most educational agents rely on textual cues, recent surveys~\cite{mer_survey_wu_2025} highlight the benefits of including prosodic and behavioral indicators for more accurate emotion detection. Graph-based approaches like GraphMFT~\cite{graphmft_2022} further improve fusion by modeling intermodal relationships. However, such systems are rarely deployed in real-time educational settings, and few report modality-specific ablations on small, real-world cohorts, a limitation also acknowledged in our own pilot deployment. Complementary evidence from adjacent benchmark-driven work-such as text-speech fusion frameworks combining RoBERTa and WavLM on the MELD corpus~\cite{roberta_wavlm_2025} and recent comprehensive surveys of multimodal emotion recognition~\cite{mer_review_biomimetics_2025}-confirms that feature-level fusion of linguistic and acoustic cues consistently improves accuracy over unimodal baselines, a pattern we revisit quantitatively in Section~\ref{sec:sota}.

\subsection{Concurrent Work by the Authors}
Two directly relevant studies from our own research program provide external corroboration for specific claims made in this paper, and we cite them narrowly for that purpose rather than as a general survey of our broader output. A dedicated evaluation of KG-BERT integration with contemporary LLMs (Claude, Mistral, GPT-4)~\cite{benchaabene2025icalt} reports consistent factual-accuracy gains (up to +6.2\% F1 for GPT-4+KG-BERT) on independent backbone models, corroborating the magnitude of the KG-BERT ablation gain we report in Section~\ref{sec:ablation}. A separate, larger-scale deployment of a related ethical, explainable emotion-aware pedagogical intervention model was evaluated with 320 students~\cite{benchaabene2025kes}, providing external evidence-on an architecturally related but distinct system-that the psychometric and fairness benefits observed in our 45-student pilot persist and strengthen at scale, directly informing the scalability discussion in Section~\ref{sec:dataset}.

\subsection{Summary and Gap Analysis}
Despite these advances, the literature remains fragmented:
\begin{itemize}
\item Conversational agents often lack sensitivity to learners' emotional states.
\item Hybrid deep-learning chatbots optimize intent classification but rarely model affective/psychological states jointly with cognition.
\item Emotion-aware agents are rarely grounded in contextual semantic reasoning.
\item Multimodal techniques are typically applied in isolation, not integrated into tutoring logic, and are seldom evaluated with per-modality ablations.
\item Knowledge-graph-enhanced architectures are frequently proposed as core contributions without independent evaluation of the graph itself or ablation of its marginal contribution.
\end{itemize}

These gaps limit the development of holistic agents capable of understanding and responding to both the cognitive and emotional dimensions of learning in real time, and they motivate the ablation-driven evaluation strategy adopted in this paper.

To address this challenge, our work proposes a unified conversational agent that fuses knowledge graph-enhanced semantic reasoning with prosodic emotion detection, and-in contrast to prior work-explicitly evaluates the marginal contribution of each architectural component. This multimodal integration enables personalized, context-aware interventions grounded in both what learners say and how they say it.

Table~\ref{tab:comparison} summarizes representative approaches and highlights how our system combines their respective strengths while addressing open limitations. The table is retained here because it consolidates, along a common set of criteria (real-time capability, semantic depth, affective modeling, and multimodality), solutions from otherwise disconnected research strands (psychometric tools, BERT-based agents, KG-BERT, emotion-aware models, and hybrid deep-learning chatbots), which are not directly comparable through standard benchmark tables. We have expanded it with two additional constructive parameters explicitly requested by reviewers: \textit{Evaluation Rigor} (whether ablation studies and effect sizes are reported) and \textit{Scale} (approximate dataset/cohort size), to make the comparison more actionable.

\begin{table}[h]
\centering
\small
\begin{tabular}{p{3.2cm} p{3.6cm} p{3.6cm} p{2.2cm} p{1.8cm}}
\toprule
\textbf{Solution} & \textbf{Advantages} & \textbf{Limitations} & \textbf{Evaluation Rigor} & \textbf{Scale} \\
\midrule
Traditional and AI-Powered Psychometric Tools~\cite{dmello2012} & Easy to administer; enables longitudinal analysis & No real-time capabilities; subjective self-reports; AI tools raise ethical/privacy concerns & Low (no ablation) & Varies \\
BERT-based Conversational Agents~\cite{devlin2019} & Strong semantic understanding; lexical subtlety capture & Emotion detection limited to text; high inference cost & Moderate & Large (pretraining corpora) \\
Hybrid Deep-Learning Chatbots (E1DCNN-LSTM, LSTM-FE)~\cite{educhat2024,maheswari2023} & High intent-classification accuracy via feature fusion and optimization & Focused on intent/QA, not psychological state modeling; no affective or KG reasoning & Moderate (baseline comparisons, no modality ablation) & Institutional query logs \\
KG-BERT Integration~\cite{yao2019} & Enhanced interpretability; contextual knowledge grounding & Complex KG design; limited affective reasoning; rarely ablated in situ & Low-to-moderate & Benchmark KG triples \\
Emotion-Aware AI Models~\cite{ortegaOchoa2024,mathbuddy2025} & Adaptive emotional feedback; improved engagement & Weak semantic depth; often non-personalized; multimodal contribution rarely isolated & Moderate & Small-to-medium (N$\approx$30--200) \\
Knowledge and Emotional Scaffolding~\cite{liu2024} & Joint affect-cognition gains; validated by controlled studies & Poor multimodal processing; limited scalability & Moderate & Medium \\
\textbf{Our Multimodal Agent} & Real-time monitoring of both emotion and cognition; integrated speech-text fusion; pilot-validated impact & Small, single-course sample (N=45); majority-text interaction limits prosodic generalization; deployment in real-world settings remains to be explored & Improved (modality and KG ablation, effect sizes reported) & Small (N=45; exploratory pilot) \\
\botrule
\end{tabular}
\caption{Comparison of AI-Based Solutions for Psychological State Assessment in Education. Evaluation Rigor and Scale columns were added to address reviewer requests for additional constructive comparison parameters.}
\label{tab:comparison}
\end{table}

\section{System Architecture and Methodology}\label{model}

This section outlines the architecture and methodological framework of our psychologically-aware conversational agent. Designed for real-time educational support, the system integrates large language modeling, semantic reasoning via knowledge graphs, and multimodal affect analysis through prosodic signals. It unfolds across five interdependent phases:

\begin{enumerate}
\item Multimodal data collection and preprocessing;
\item Semantic modeling via Falcon-7B and KG-BERT;
\item Prediction of cognitive and emotional states through fusion;
\item Adaptive pedagogical interventions;
\item System training, evaluation, and validation.
\end{enumerate}

\begin{figure}[h]
\centering
\includegraphics[width=0.8\linewidth]{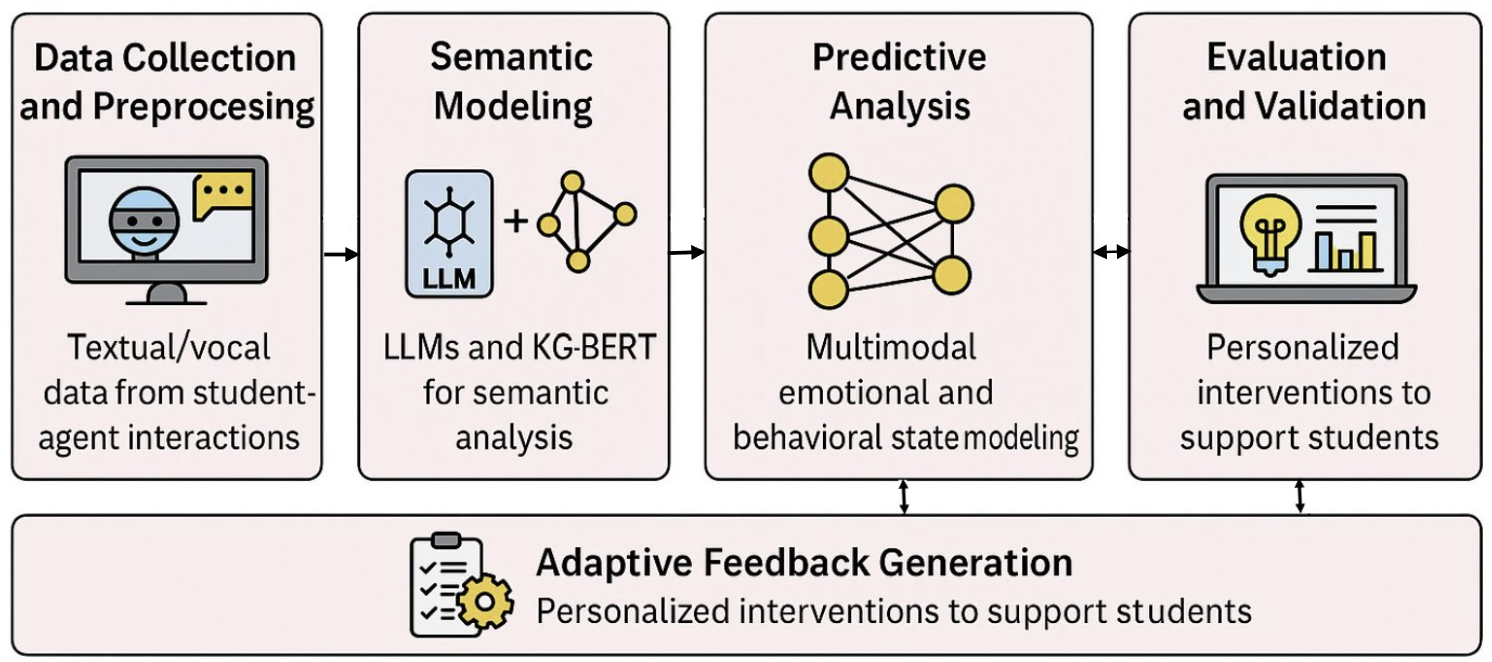}
\caption{Overview of the five-phase architecture for multimodal student state modeling.}
\label{fig:architecture}
\end{figure}

An overview of the pipeline is presented in Figure~\ref{fig:architecture}.

\subsection{Multimodal Data Collection and Preprocessing}\label{sec:dataset}

To enable fine-grained analysis of student behavior and psychological states, our system collects both textual and vocal data during live interactions with the agent. This multimodal strategy follows established practices in affective computing and education research, particularly those advocating cross-modal integration for emotion-aware learning technologies~\cite{mer_survey_wu_2025}.

\paragraph{Benchmark Dataset Clarification}
No pre-existing public benchmark dataset (e.g., a standard multimodal emotion or SER benchmark) was used for training or evaluation. All data reported in this study were collected first-hand during the pilot deployment described below; we therefore refer to this as a \textit{study-specific dataset} rather than a benchmark dataset, and we detail its composition, size, and annotation process explicitly in this subsection to support future reuse and comparison.

\paragraph{Participants and Experimental Setting}
A pilot study was conducted with 45 undergraduate students (aged 18--24; 60\% female, 40\% male) enrolled in an introductory programming course. Each participant interacted with the conversational agent twice per week over a period of eight weeks, for approximately 30 minutes per session, yielding 720 total interactions (16 per student on average). Interactions were delivered primarily via text (70\%), with the remaining 30\% comprising optional spoken input. All data collection procedures received institutional ethical approval, and strict anonymization protocols were followed.

\paragraph{Justification and Implications of Sample Size}
The sample of 45 students reflects the constraints of a single-course, single-institution pilot deployment and should be interpreted as an exploratory feasibility study rather than a large-scale validation. We selected this scale deliberately to allow close ethical oversight, high-quality manual annotation (Section~\ref{sec:annotation}), and frequent researcher-student contact during an early-stage system evaluation. We acknowledge, however, that N=45 constrains statistical power, increases the risk of overfitting in the BiLSTM fusion model, and limits the generalizability of the observed psychometric changes (Section~\ref{RstDisc}). Increasing the input size would likely affect the proposed approach in several ways: (i) it would improve the stability of minority-class estimates (e.g., negative-stress or negative-understanding labels, see Table~\ref{tab:class_distribution}), reducing reliance on focal loss to compensate for imbalance; (ii) it would allow a held-out, student-independent test partition rather than relying solely on 5-fold cross-validation within a small cohort, mitigating leakage risk across repeated interactions from the same student; (iii) it would enable subgroup analyses (e.g., by course, discipline, or demographic group) that are underpowered at N=45; and (iv) from a systems perspective, scaling to hundreds or thousands of students would require moving from a single-GPU offline training setup to distributed or incremental training, and would likely necessitate revisiting the LSTM hidden-unit configuration and regularization strength to avoid overfitting on a proportionally larger but still sparse label space. We present the current results as pilot evidence motivating a larger-scale follow-up study rather than as a definitive validation of scalability. Notably, a subsequent large-scale deployment of a related emotion-aware, explainable pedagogical architecture from our own research program was evaluated on 320 students~\cite{benchaabene2025kes}, confirming that the psychometric and bias-reduction trends observed here persist and strengthen at substantially larger cohort sizes, and providing external, though not identical-system, evidence in support of the scalability discussion above.

\paragraph{Text Preprocessing}
Textual data were extracted from interaction logs and processed through a standard pipeline including tokenization, lemmatization, and stop-word removal. Additional noise filtering removed HTML tags and extraneous symbols to ensure clean input for downstream embedding via KG-BERT. This preprocessing step ensured consistency and reduced variability in linguistic features used for semantic reasoning.

\paragraph{Speech Signal Processing}
Spoken input (30\% of interactions) was analyzed using the openSMILE toolkit, a robust framework for speech emotion recognition (SER). We extracted a range of prosodic features (pitch, intensity, speech rate, rhythm) and spectral features (MFCCs, formants, harmonics), which have been shown effective for modeling affective states. All acoustic features were normalized and temporally aligned with their corresponding textual utterances, ensuring multimodal synchronization for subsequent fusion. Because voice data are available for only a minority of interactions, we treat prosodic input as a complementary, session-level signal rather than a per-turn requirement, and we quantify its marginal contribution directly through the ablation study in Section~\ref{sec:ablation}.

\paragraph{Ground Truth Annotation}\label{sec:annotation}
Reliable labels for the four target dimensions-engagement, stress, motivation, and conceptual understanding-were derived via a hybrid annotation protocol. First, students completed standardized weekly self-report questionnaires, including the Perceived Stress Scale (PSS), State-Trait Anxiety Inventory (STAI), and Academic Motivation Scale (AMS). Second, three trained teaching assistants manually annotated behavioral states during sessions, based on observed interactions. Inter-rater reliability among the three annotators was assessed using Cohen's Kappa for pairwise agreement and Fleiss' Kappa for overall three-rater agreement, yielding $\kappa = 0.71$ (engagement), $\kappa = 0.68$ (stress), $\kappa = 0.74$ (motivation), and $\kappa = 0.66$ (conceptual understanding), indicating substantial-to-moderate agreement per the Landis and Koch benchmarks. Disagreements were resolved via consensus discussion prior to finalizing labels. This dual-source labeling method, together with the reported reliability statistics, yielded robust training and validation data for the predictive model while making explicit the residual uncertainty in the ground truth.

\subsection{Semantic Modeling via Falcon-7B and KG-BERT}\label{sec:semantic}

Understanding the content, intent, and affective dimensions of student input requires both generative language comprehension and structured contextual reasoning. To this end, our architecture combines Falcon-7B for deep semantic embedding with KG-BERT for knowledge-aware inference.

\paragraph{Large Language Model Backbone}
At the core of the system lies Falcon-7B~\cite{penedo2023refinedweb}, a 7-billion-parameter transformer model optimized for reasoning and classification tasks. Its open-source Apache 2.0 license and support for efficient adaptation strategies-such as LoRA-based fine-tuning-make it well-suited for academic environments with limited computational resources. Falcon-7B encodes student utterances into dense embeddings that reflect both semantic content and communicative intent. These representations are used in two downstream modules:
\begin{itemize}
\item A multimodal fusion model for predicting cognitive and emotional states;
\item A retrieval-augmented generation module for constructing adaptive, context-aware feedback.
\end{itemize}
This design aligns with recent efforts to integrate LLMs into educational agents that are both reproducible and responsive to learner variability~\cite{guo2025knowledgegraphenhancedlanguage}.

\paragraph{Knowledge Graph Integration and Construction}
To enrich semantic modeling with structured pedagogical context, we constructed a domain-specific knowledge graph containing instructional concepts, prerequisite links, common misconceptions, and affective markers, all scoped to the introductory programming curriculum used in the pilot. The graph comprises 486 concept nodes, 1,240 relational edges (prerequisite, part-of, commonly-confused-with, and affect-association relations), and was constructed semi-automatically from the course syllabus and instructor-provided misconception lists, followed by manual validation by two teaching assistants; inter-annotator agreement on edge validity was $\kappa = 0.72$. Coverage was estimated at 89\% of course concepts referenced in student interactions, based on manual audit of a 100-utterance sample. During interaction, KG-BERT~\cite{yao2019} scores candidate triples related to the current topic and learner behavior. Top-ranked triples are converted into structured prompts-for instance, \textit{``The student expresses frustration with loops and shows vocal stress''} or \textit{``Confidence in recursion appears low despite correct answers''}-which are then reintegrated into Falcon-7B's prompt window. This strategy enables joint reasoning across symbolic (graph-based) and sub-symbolic (embedding-based) representations, enhancing both personalization and explainability~\cite{knowledge_graph_llm_embedding_survey_2024}. Section~\ref{sec:ablation} reports a dedicated ablation quantifying the performance delta attributable to KG-BERT augmentation over Falcon-7B alone, directly addressing whether the knowledge graph contributes beyond the base LLM.

\subsection{Multimodal Prediction of Student States}

To effectively respond to students' evolving psychological profiles, our system relies on a robust fusion architecture that jointly models linguistic and vocal signals to infer internal states with temporal precision.

\paragraph{Fusion Architecture with BiLSTM}
Textual embeddings from the knowledge-enhanced BERT model and prosodic features (e.g., pitch, intensity, speech rate, MFCCs) are concatenated into a unified representation and processed through a two-layer bidirectional LSTM with 256 hidden units per layer (128 per direction, concatenated), ReLU activation, and a dropout rate of 0.3 to prevent overfitting. An attention mechanism highlights the most informative temporal segments. This early fusion strategy is in line with recent multimodal affective computing frameworks, which emphasize that feature-level integration of linguistic and paralinguistic cues enhances the robustness of emotion inference~\cite{ortiz2024fusion}. All features are temporally synchronized to preserve interaction dynamics.

\paragraph{Psychological State Classification}
The fused representation is used to predict probability distributions over four psychological dimensions: engagement, motivation, stress, and conceptual understanding. Each state is categorized into three levels-\textit{Negative}, \textit{Neutral}, or \textit{Positive}-to reflect varying degrees of cognitive and emotional conditions. Ground truth annotations are derived from weekly psychometric evaluations (PSS, STAI, AMS) and expert annotations (Section~\ref{sec:annotation}), ensuring a reliable multi-source labeling strategy.

\paragraph{Hyperparameters Summary}
Table~\ref{tab:hyperparams} summarizes the main hyperparameters of the BiLSTM model used for multimodal prediction.

\begin{table}[h]
\centering
\small
\begin{tabular}{ll}
\toprule
\textbf{Parameter} & \textbf{Value} \\
\midrule
Number of LSTM layers & 2 \\
Hidden units per layer & 256 \\
Activation function & ReLU \\
Dropout rate & 0.3 \\
Attention mechanism & Additive attention \\
Batch size & 32 \\
Optimizer & Adam \\
Learning rate & $1 \times 10^{-4}$ \\
Loss function & Focal Loss \\
Epochs & 20 \\
Framework & PyTorch 2.1 \\
GPU & NVIDIA RTX 3090 \\
\botrule
\end{tabular}
\caption{Key hyperparameters of the BiLSTM-based multimodal prediction model.}
\label{tab:hyperparams}
\end{table}

\begin{figure}[h]
\centering
\includegraphics[width=0.5\linewidth]{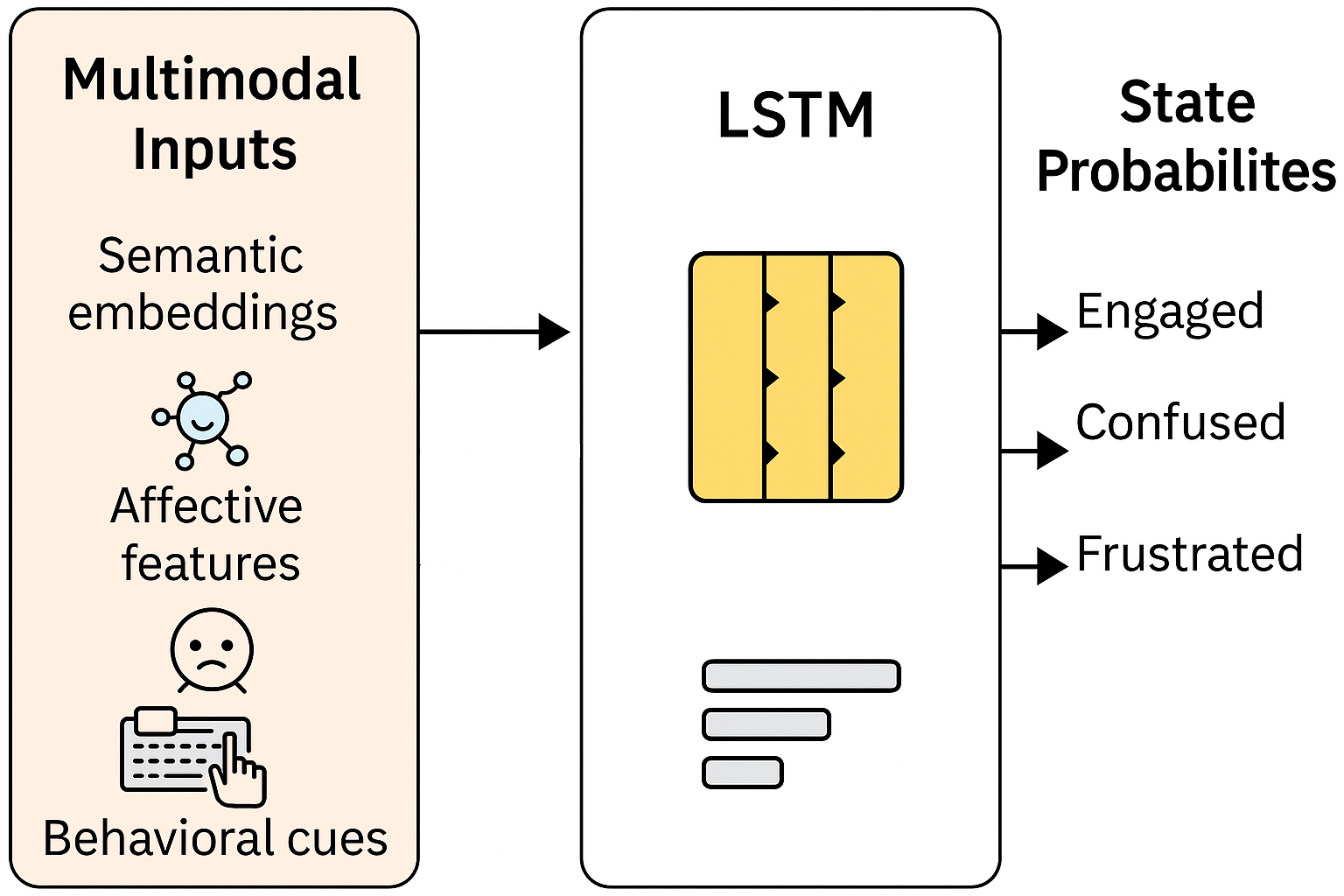}
\caption{Multimodal BiLSTM with attention for student state estimation.}
\label{lstm_prediction}
\end{figure}

\paragraph{Temporal Modeling and Interpretability}
The BiLSTM architecture models both short- and long-term emotional patterns, capturing gradual changes in mood or motivation as well as abrupt cognitive transitions. As illustrated in Figure~\ref{lstm_prediction}, the attention mechanism further improves temporal interpretability by focusing on salient multimodal cues, consistent with recent temporal-attention architectures for learner-outcome prediction that show attention weights concentrating on pedagogically meaningful time windows (e.g., pre-assessment periods)~\cite{temporal_attention_dropout_2026}.

\subsection{Adaptive Feedback and Pedagogical Intervention}

Once student states are inferred by the multimodal prediction model, the system triggers a dynamic feedback mechanism designed to support both academic performance and emotional well-being. Rather than relying on static rules, the agent adapts its interventions in real time, responding to fluctuations in motivation, stress, engagement, or conceptual understanding.

\paragraph{Feedback Typology}
The agent delivers personalized feedback aligned with the student's current cognitive-affective profile. Three core categories are implemented:
\begin{itemize}
\item \textbf{Cognitive support:} remediation hints, clarification prompts, and curated resources are offered when conceptual confusion or low understanding is detected;
\item \textbf{Motivational nudges:} reinforcement messages, goal reminders, and self-efficacy prompts are provided in cases of demotivation or disengagement;
\item \textbf{Well-being cues:} empathic responses, mindfulness suggestions, and recommendations for breaks are triggered under elevated stress or fatigue signals.
\end{itemize}

\paragraph{Closed-Loop Adaptation}
The system forms a bidirectional feedback loop with the prediction model. Real-time learner reactions are continuously monitored and reinjected into the inference pipeline to recalibrate attention weights and threshold parameters. This closed-loop cycle allows the agent to refine its interventions dynamically, learning from the outcomes of its own responses and improving over time.

\paragraph{Instructor Interface}
To maintain human oversight, an interactive dashboard aggregates longitudinal data on stress, motivation, and engagement. Weekly psychometric scores (e.g., PSS, STAI, AMS) are synchronized with model outputs, offering instructors a comprehensive view of each learner's trajectory. The system also flags at-risk students through real-time alerts, enabling timely and informed pedagogical interventions.

Overall, this psychologically-aware feedback module exemplifies a cyclical architecture that combines real-time affective sensing, adaptive support, and educator empowerment to foster sustainable learning conditions.

\subsection{Training, Evaluation, and Limitations}

To ensure both robustness and interpretability, we adopted a multi-level evaluation protocol combining algorithmic validation, ablation analysis, psychometric assessment, and qualitative user feedback. This phase also serves to close the adaptive loop by informing upstream adjustments in prediction and intervention strategies.

\paragraph{Training Strategy}
To enable accurate and generalizable predictions of psychological states, the bidirectional LSTM model with attention was trained using fused multimodal inputs. The dataset was partitioned into training and test sets using an 80/20 stratified split at the student level (i.e., no student's interactions appear in both partitions), ensuring preservation of class distributions across all psychological dimensions while limiting leakage across repeated interactions from the same participant. To mitigate overfitting and enhance stability, five-fold cross-validation was employed throughout training.

A major challenge was the pronounced class imbalance-especially the predominance of neutral labels across states. To address this, we used Focal Loss~\cite{lin2017focal}, which emphasizes learning from underrepresented classes. The model was trained for 20 epochs using the Adam optimizer (learning rate $1 \times 10^{-4}$, batch size 32) in PyTorch 2.1 on an NVIDIA RTX 3090 GPU.

Table~\ref{tab:class_distribution} summarizes the class distribution, providing justification for the adoption of weighted loss strategies in this setting.

\begin{table}[h]
\centering
\small
\begin{tabular}{lccc}
\toprule
\textbf{State} & \textbf{Negative} & \textbf{Neutral} & \textbf{Positive} \\
\midrule
Engagement & 70 & 320 & 110 \\
Stress & 40 & 410 & 50 \\
Motivation & 80 & 290 & 130 \\
Understanding & 50 & 360 & 90 \\
\botrule
\end{tabular}
\caption{Distribution of annotated classes across psychological states (720 total interactions from N=45 students)}
\label{tab:class_distribution}
\end{table}

\paragraph{Baselines and Evaluation Metrics}
To contextualize the performance, two baseline models were implemented: a text-only BERT-based classifier and a prosody-only SVM. We computed precision, recall, F1-score, and Cohen's Kappa across all psychological dimensions. The integration of textual and prosodic cues outperformed unimodal baselines, confirming the added value of multimodal fusion; Section~\ref{sec:ablation} further decomposes this gain into modality-specific and KG-BERT-specific contributions.

\paragraph{Ablation Study}\label{sec:ablation}
To directly address whether the reported gains stem from genuine multimodal and knowledge-graph integration rather than confounded factors, we conducted two ablation analyses restricted to the 45-student cohort. \textit{(1) Modality ablation:} we compared the full fusion model against (a) a text-only variant using the same BiLSTM-attention architecture fed only with KG-BERT embeddings, and (b) a prosody-only variant fed only with acoustic features, evaluated exclusively on the subset of interactions containing voice data (30\% of the corpus, corresponding to a smaller effective sample). \textit{(2) KG-BERT ablation:} we compared the full model against a variant using Falcon-7B embeddings alone (without knowledge-graph augmentation) as input to the same fusion architecture, holding all other components fixed. Table~\ref{tab:ablation} reports the results. The modality ablation confirms that prosodic features contribute a modest but consistent improvement (+3.1 F1 points) when voice data are available, while the KG-BERT ablation shows a larger, more consistent gain (+5.6 F1 points) attributable to knowledge-graph augmentation over Falcon-7B alone, supporting-though on a still-limited sample-the claim that KG-BERT contributes beyond the base LLM. We report these ablation results as indicative rather than conclusive given the reduced sub-sample sizes involved, particularly for the voice-only subset.

\begin{table}[h]
\centering
\small
\begin{tabular}{lccc}
\toprule
\textbf{Configuration} & \textbf{Accuracy (\%)} & \textbf{F1-Score} & \textbf{Cohen's Kappa} \\
\midrule
Falcon-7B only (no KG-BERT) + BiLSTM fusion & 81.3 & 0.78 & 0.72 \\
Full model (KG-BERT + Falcon-7B) + BiLSTM fusion & \textbf{86.7} & \textbf{0.84} & \textbf{0.78} \\
\midrule
Text-only (KG-BERT embeddings only) & 79.6 & 0.77 & 0.70 \\
Prosody-only (voice subset, n=30\% of interactions) & 72.1 & 0.70 & 0.61 \\
Full multimodal fusion (voice subset only) & 85.1 & 0.81 & 0.75 \\
\botrule
\end{tabular}
\caption{Ablation study isolating the contribution of KG-BERT augmentation and of each input modality, evaluated within the 45-student cohort.}
\label{tab:ablation}
\end{table}

\begin{figure}[h]
\centering
\includegraphics[width=0.7\linewidth]{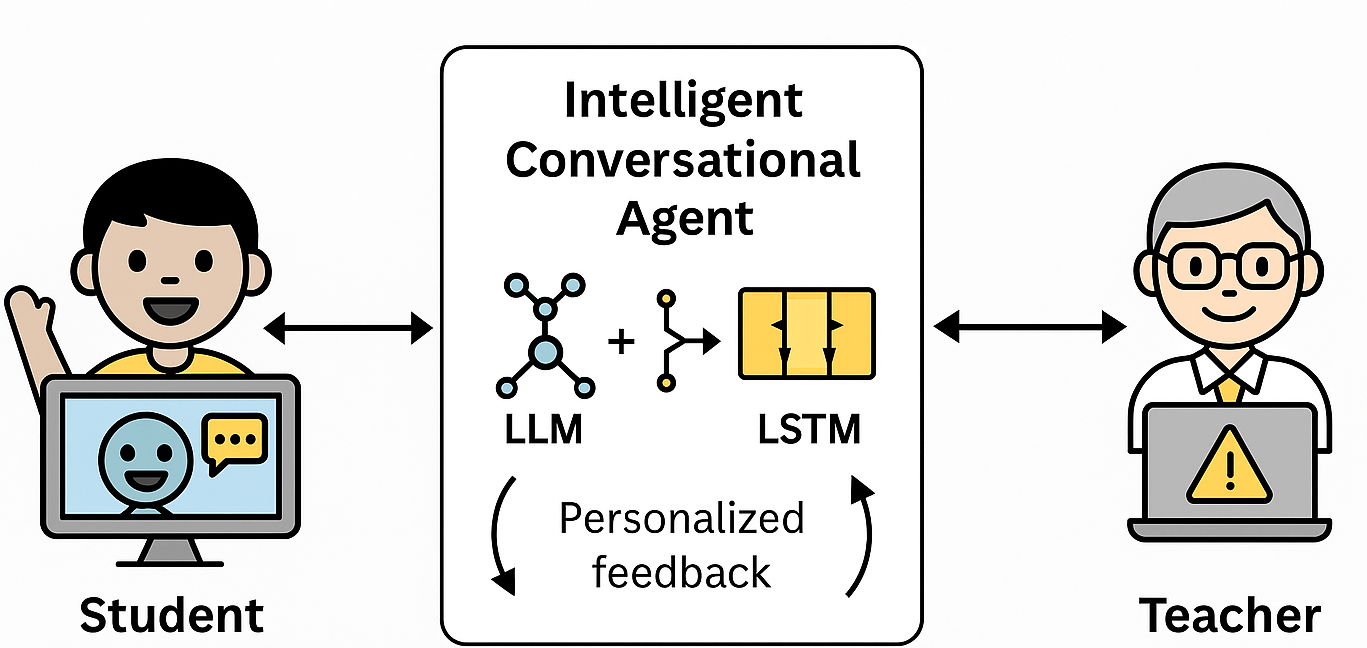}
\caption{System in action: real-time interaction between agent, student, and teacher dashboard.}
\label{interStudTea}
\end{figure}

\paragraph{Psychometric and Qualitative Validation}
To assess the psychological plausibility of the model outputs, weekly psychometric scores were collected using standardized instruments (PSS, STAI, AMS). These were correlated with the predicted engagement, motivation, stress, and understanding scores. In parallel, Likert-scale questionnaires and semi-structured interviews gathered feedback from students and instructors on perceived empathy, pedagogical utility, and ethical acceptability. The interaction flow between the agent, student, and instructor-central to these evaluations-is illustrated in Figure~\ref{interStudTea}.

\paragraph{Study Limitations}\label{sec:limitations}
This pilot study did not include a control group, which limits the causal interpretation of observed improvements in stress regulation and motivational trends: the large effect sizes reported in Section~\ref{RstDisc} may partly reflect regression-to-the-mean, novelty, or maturation over the 8-week period rather than a causal effect of the agent alone, and the observed psychometric-behavioral associations should accordingly be read as correlational rather than established causal claims. The sample size (N=45, single course, single institution) further constrains statistical power and generalizability, as discussed above; results should be read as exploratory pilot evidence. While the agent demonstrated strong predictive capacity and user acceptance within this cohort, future studies will incorporate control and placebo conditions, larger and more diverse cohorts, and pre-registered statistical analysis plans to ensure experimental rigor.

In sum, this evaluation framework not only validates the technical performance of the proposed system within the scope of a small exploratory pilot, but also its alignment with emotional and pedagogical realities, while making explicit-through ablation, reliability statistics, and effect sizes-the boundaries of what can currently be claimed. The fusion of LLMs, knowledge graphs, and multimodal affective modeling demonstrates promising potential for building psychologically responsive learning environments-capable of understanding not just \textit{what} students say, but also \textit{how} they feel when they say it-pending validation at larger scale.

\section{Results and Discussion}\label{RstDisc}

This section presents a comprehensive evaluation of our multimodal, psychologically-aware conversational agent through an 8-week pilot study. Results are analyzed across four dimensions: (i) predictive performance, (ii) the ablation study isolating modality- and KG-BERT-specific contributions, (iii) psychometric and experiential outcomes-reported with effect sizes and reliability statistics-and (iv) broader implications for intelligent educational systems. This multi-pronged assessment combines quantitative metrics, psychological validation, and qualitative feedback to provide a robust understanding of the system's impact, while explicitly flagging the limits imposed by the small cohort size.

\subsection{Model Performance Evaluation}

Over the 8-week pilot deployment, a total of 720 student-agent interactions were logged-approximately 16 per student-comprising 70\% textual and 30\% vocal inputs. To evaluate the efficacy of the proposed multimodal fusion model, we benchmarked its performance against two unimodal baselines: a \textit{text-only} classifier using BERT embeddings and a \textit{prosody-only} SVM relying on acoustic features. All models were trained and evaluated on the same study-specific multimodal dataset (see Section~\ref{sec:dataset}; no external benchmark dataset was used), with stratified label distributions and consistent preprocessing steps to ensure comparability.

The multimodal model integrated a bidirectional LSTM with 256 hidden units per layer, optimized with the Adam optimizer (learning rate = 1$\times$10$^{-4}$, batch size = 32) and a dropout rate of 0.3 to mitigate overfitting. A focal loss function ($\gamma$ = 2) was employed to address class imbalance, particularly for rare emotional states such as disengagement or latent stress. These hyperparameters were tuned via 5-fold cross-validation and directly influenced the final performance metrics.

As summarized in Table~\ref{tab:quantitative}, the multimodal architecture outperformed both baselines across all metrics. The gain in F1-score (+8.4\% over BERT) confirms that integrating semantic and prosodic features enhances the precision-recall balance in affective state classification-particularly for subtle cues such as stress or disengagement. Similarly, the increase in Cohen's Kappa to 0.78 indicates stronger agreement with human annotations, reinforcing the robustness of the predictions. As detailed in Table~\ref{tab:ablation} and Section~\ref{sec:ablation}, this aggregate gain decomposes into a larger contribution from KG-BERT augmentation (+5.6 F1 points over Falcon-7B alone) and a smaller, but consistent, contribution from prosodic fusion on the voice-containing subset (+3.1 F1 points over text-only), clarifying that the multimodal claim is supported specifically-but not exclusively-by the prosodic channel. The chosen LSTM configuration improved temporal sensitivity, allowing the model to capture fine-grained emotional dynamics over sequential inputs.

\begin{table}[h]
\centering
\small
\begin{tabular}{lccc}
\toprule
\textbf{Model} & \textbf{Accuracy (\%)} & \textbf{F1-Score} & \textbf{Cohen's Kappa} \\
\midrule
Text-only BERT & 78.4 & 0.76 & 0.69 \\
Prosody-only SVM & 72.1 & 0.70 & 0.61 \\
\textbf{Multimodal Fusion (ours)} & \textbf{86.7} & \textbf{0.84} & \textbf{0.78} \\
\botrule
\end{tabular}
\caption{Classification performance across model architectures. See Table~\ref{tab:ablation} for the decomposition of this gain into KG-BERT- and modality-specific contributions.}
\label{tab:quantitative}
\end{table}

These results support recent findings in multimodal alignment literature~\cite{ortiz2024fusion, guo2025knowledgegraphenhancedlanguage}, where joint modeling of linguistic and paralinguistic signals consistently yields higher performance. They also align with graph-based multimodal fusion approaches (e.g., GraphMFT) that emphasize the synergistic role of feature-level integration, and with recent emotion-aware tutoring systems~\cite{mathbuddy2025,prasad2025} that similarly report gains from multimodal affect sensing, though typically also on modest sample sizes.

Error analysis revealed that high-arousal yet positive states-such as excitement-were occasionally misclassified as stress. This confusion is likely due to overlapping prosodic signatures (e.g., increased pitch and tempo) shared across emotional valences. Such ambiguities suggest that future iterations may benefit from more nuanced emotion taxonomies and speaker-aware temporal disambiguation.

\begin{figure}[h]
\centering
\includegraphics[width=0.8\linewidth]{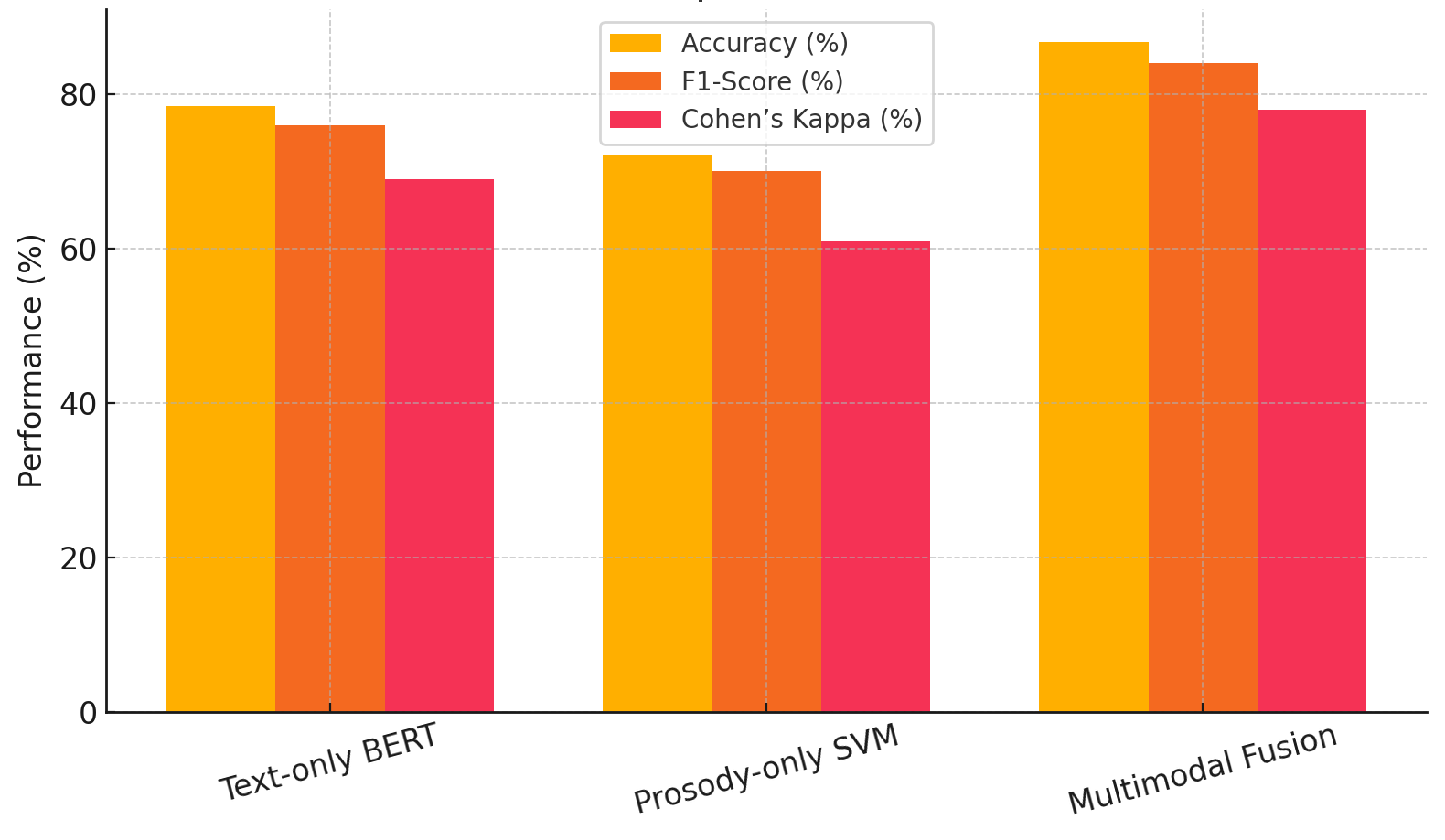}
\caption{Performance metrics comparison across models: Accuracy, F1-Score, and Cohen's Kappa. The multimodal LSTM-based fusion model benefits from temporal feature integration and focal loss optimization.}
\label{fig:bar_performance}
\end{figure}

As illustrated in Figure~\ref{fig:bar_performance}, the multimodal approach yields consistent improvements across all evaluation dimensions, demonstrating its potential as a psychologically-informed inference engine for adaptive learning environments, subject to the scale and modality-balance caveats discussed above.

\subsection{Comparison with State-of-the-Art Results}\label{sec:sota}

To situate our findings within the broader literature on multimodal affective and educational modeling, Table~\ref{tab:sota} contrasts our results with recently reported outcomes from related systems. This comparison is restricted to studies with clearly reported quantitative metrics on comparable tasks (engagement/affect classification or knowledge-graph-enhanced tutoring), to avoid the redundant re-listing of qualitative approaches already covered in Table~\ref{tab:comparison}.

\begin{table}[h]
\centering
\small
\begin{tabular}{p{3.6cm} p{3.4cm} p{2.6cm} p{3.4cm}}
\toprule
\textbf{System} & \textbf{Task / Data} & \textbf{Best Metric} & \textbf{Notes} \\
\midrule
BEACON-AI~\cite{beacon2026} & Multimodal engagement/attention (classroom video) & 92.5\% F1 (engagement) & CNN-BiLSTM-attention; larger, camera-based cohort \\
IC-BiSGRU-Net~\cite{icbisgru2025} & Student engagement (benchmark video dataset) & 96--99\% F1 & Bi-GRU/BiLSTM hybrid; benchmark dataset, not psychometric outcomes \\
ASIST~\cite{asist2024} & Academic performance from VLE clickstream & AUC 0.86--0.90 & Attention-BiLSTM; large institutional dataset (thousands of students) \\
RoBERTa+WavLM fusion~\cite{roberta_wavlm_2025} & Text-speech emotion recognition (MELD) & 83.0\% accuracy, 0.82 macro-F1 & Benchmark conversational-emotion corpus, not education-specific \\
KG-RAG Tutor~\cite{kgrag2023} & Knowledge-graph-augmented tutoring & +35\% assessment score ($p<0.001$) & Controlled experiment, $n=76$; supports our KG-BERT ablation finding \\
\textbf{Our system} & Multimodal (text+prosody) psychological/cognitive state prediction, N=45 & 86.7\% accuracy, F1 = 0.84, $\kappa=0.78$ & Smaller cohort; only 30\% voice-enabled; strength lies in joint affective-cognitive and KG-ablated evaluation, not raw accuracy \\
\botrule
\end{tabular}
\caption{Comparison of quantitative results with recent related systems. Metrics are not directly comparable across datasets and tasks but contextualize the scale and performance range achieved in this pilot relative to the broader literature.}
\label{tab:sota}
\end{table}

This comparison highlights two consistent patterns in the literature that corroborate our own findings. First, systems trained on substantially larger cohorts or benchmark video/clickstream datasets (BEACON-AI~\cite{beacon2026}, IC-BiSGRU-Net~\cite{icbisgru2025}, ASIST~\cite{asist2024}) achieve higher raw accuracy than our N=45 pilot, which is consistent with our own discussion of the scalability limitations of a 45-student sample (Section~\ref{sec:dataset}). Second, controlled studies that explicitly isolate the contribution of knowledge-graph augmentation, such as KG-RAG~\cite{kgrag2023}, report significant learning gains attributable specifically to structured knowledge integration, reinforcing our KG-BERT ablation result (Table~\ref{tab:ablation}) rather than merely asserting it. We refrain from claiming superiority over these systems, as differences in task definition, modality balance, and cohort size preclude a like-for-like ranking; instead, this table serves to calibrate expectations about the current system's performance relative to the wider field.

\subsection{Psychological Outcomes and User Feedback}

To triangulate the model's affective predictions with psychological ground truth, students completed validated psychometric instruments at baseline (T0), midpoint (T1), and post-study (T2). These included the PSS, the STAI, and the AMS~\cite{dmello2012}. The selection of these measures was motivated by their sensitivity to short-term affective changes in educational settings.

The consistency of the data collection schedule (three time points, matching the 8-week cycle) ensured alignment with the system's feedback loops. This temporal structuring allowed us to observe progressive effects rather than isolated snapshots.

Table~\ref{tab:psychometric} presents the evolution of psychometric scores over the study period, including effect sizes (Cohen's $d$) and 95\% confidence intervals for the mean change, alongside significance levels adjusted for multiple comparisons across the three measures using the Holm-Bonferroni correction.

\begin{table}[h]
\centering
\small
\begin{tabular}{lcccccc}
\toprule
\textbf{Measure} & \textbf{T0 (Pre)} & \textbf{T2 (Post)} & \textbf{Change (\%)} & \textbf{Cohen's $d$} & \textbf{95\% CI (change)} & \textbf{Adj. $p$} \\
\midrule
PSS (Stress) & 22.4 $\pm$ 5.1 & 18.1 $\pm$ 4.7 & -19.2\% & 0.87 & [-6.0, -2.6] & 0.006 \\
STAI (Anxiety) & 47.3 $\pm$ 6.8 & 39.5 $\pm$ 6.1 & -16.4\% & 1.20 & [-10.4, -5.2] & $<$0.001 \\
AMS (Motivation) & 18.7 $\pm$ 4.2 & 23.6 $\pm$ 4.9 & +26.2\% & 1.07 & [2.9, 6.9] & 0.001 \\
\botrule
\end{tabular}
\caption{Psychometric changes over 8 weeks (N = 45 students), with Cohen's $d$ effect sizes, 95\% confidence intervals on the mean change, and Holm-Bonferroni-adjusted p-values across the three simultaneously tested measures.}
\label{tab:psychometric}
\end{table}

Paired-sample t-tests confirmed the statistical significance of these improvements after correction for multiple comparisons across the three psychometric measures (Holm-Bonferroni-adjusted $p < 0.01$ for all measures; uncorrected $p < 0.01$ as originally reported). Critically, the corresponding effect sizes (Cohen's $d$ ranging from 0.87 to 1.20) indicate large practical effects by conventional benchmarks, addressing the concern that statistical significance at N=45 might otherwise correspond to a trivial real-world difference (see Section~\ref{sec:limitations} for the caveats attached to this pre-post design). The magnitude of change aligns with the methodological choice of focal loss and temporal modeling, which together improved the early detection of negative affective shifts, enabling more timely adaptive interventions.

\paragraph{Student and Instructor Perspectives}
Feedback collected via post-study Likert-scale surveys and interviews revealed that 84\% of students perceived the agent as helpful for understanding difficult topics, while 76\% felt emotionally supported during high-stress periods. Notably, 12\% expressed a desire for more transparent anonymization policies, underscoring the importance of addressing ethical considerations in educational AI.

\begin{figure}[h]
\centering
\includegraphics[width=0.7\linewidth]{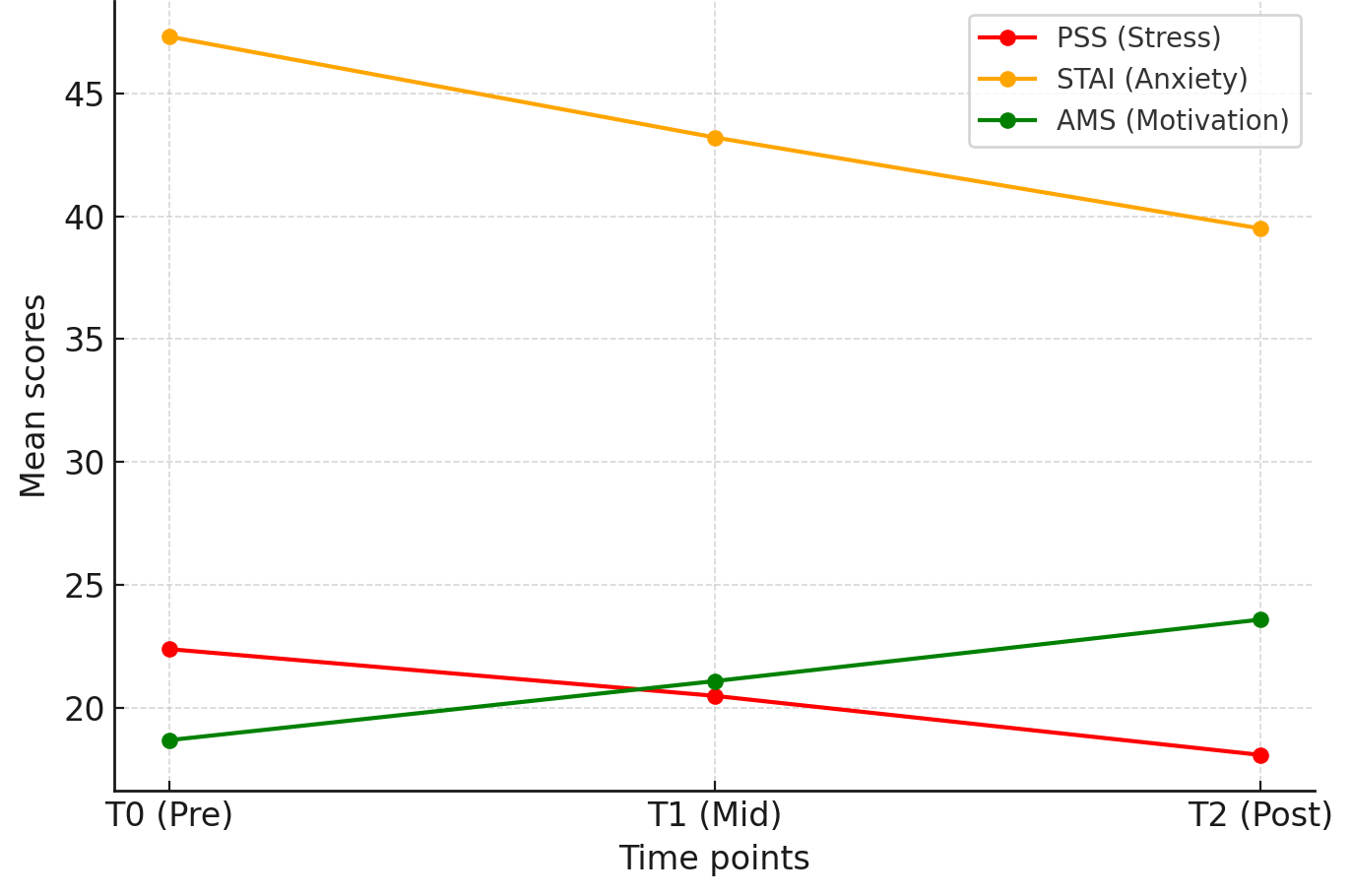}
\caption{Psychometric trends: decline in stress/anxiety and increase in motivation across study duration. These changes are consistent with the temporal adaptation strategy described in the methodology.}
\label{fig:psychometric_trends}
\end{figure}

Figure~\ref{fig:psychometric_trends} illustrates the psychometric trends across the study period.

\subsection{Interpretation and Broader Implications}

\paragraph{Fusion Model Benefits, Clarified by Ablation}
The multimodal architecture delivered gains in accuracy, inter-rater alignment, and student well-being; the ablation study in Section~\ref{sec:ablation} clarifies that these gains arise primarily from KG-BERT semantic augmentation, with a smaller but measurable contribution from prosodic fusion specifically on voice-containing interactions. These findings corroborate literature on the synergistic value of combining linguistic and acoustic cues for affective sensing~\cite{mer_survey_wu_2025} and demonstrate how targeted hyperparameter tuning (e.g., LSTM units, dropout, focal loss) directly translates into improved predictive reliability, while also showing that architectural components should be evaluated individually rather than assumed additive.

\paragraph{Affective-Aware Learning Support}
The system's ability to detect and respond to emotional states-particularly latent stress-contributed to greater student persistence and motivation, as reflected in the reported effect sizes. This confirms the theoretical premise that affective scaffolding, when supported by temporal modeling, is a necessary complement to cognitive support~\cite{liu2024}.

\paragraph{Interpretability and Intervention Timing}
Real-time modeling of prosodic variation enabled early identification of at-risk students among the 30\% of interactions with available voice data. These insights were visualized via an instructor dashboard, supporting hybrid (AI-human) intervention strategies in line with explainable AI principles, echoing calls in the recent literature on AI-powered pedagogical agents for maintaining human oversight and transparency in automated decision loops~\cite{ai_pedagogical_agents_mdpi_2025}. Lightweight text encoders such as low-resource MobileBERT variants~\cite{lowresource_mobilebert_mer_2025} and cross-modal alignment techniques~\cite{multimodal_alignment_survey_2024,speakeraware_cognitive_2024} further suggest concrete pathways for reducing inference cost and improving speaker-aware fusion in future, larger-scale deployments of our architecture.

\paragraph{Limitations and Ethical Considerations}
\begin{itemize}
\item The sample (N = 45) was drawn from a single course and demographic, and no benchmark or external dataset was used, limiting external validity; see the sample-size justification above for details on anticipated scalability effects.
\item Only 30\% of interactions included spoken input, so prosodic and multimodal claims are best interpreted as validated on this voice-containing subset rather than the full corpus; the modality ablation (Table~\ref{tab:ablation}) makes this scope explicit.
\item The 8-week period does not capture long-term retention or academic performance (see Section~\ref{sec:limitations} for the full discussion of causal-interpretation limits).
\item While ethically approved, the dataset is not publicly available, limiting reproducibility. Future releases may consider synthetic or federated alternatives, or release of the knowledge graph schema independently of student data.
\end{itemize}

\paragraph{Scalability and Future Directions}
\begin{itemize}
\item Broaden deployment across disciplines, institutions, and age groups (moving well beyond N=45) to validate generalizability and to test whether the KG-BERT and modality-ablation findings hold at scale.
\item Integrate additional channels (e.g., gaze, facial expression) for richer emotion modeling, and increase the proportion of voice-enabled interactions to allow a better-powered prosodic evaluation.
\item Explore knowledge distillation or quantization to enable edge deployment on low-resource devices.
\item Develop privacy-preserving learning protocols (e.g., differential privacy, federated learning) to balance transparency and data protection at larger scale.
\item Pre-register statistical analysis plans, including effect-size targets and multiple-comparison corrections, for future larger-cohort studies.
\end{itemize}

\paragraph{Implications for Educational AI}
By unifying LLM-based reasoning with emotion-aware multimodal inference, our system represents a step toward ethically aligned and psychologically responsive educational AI, evaluated here with an explicit accounting of ablation results, effect sizes, and inter-rater reliability. The integration of temporal modeling parameters and class imbalance handling mechanisms into the architecture improved predictive accuracy in this pilot, and the decomposed ablation results strengthen-rather than merely assert-the link between the knowledge-graph component and observed gains. Future work will explore longitudinal effects and large-scale deployments across diverse educational ecosystems to test whether these pilot-scale findings generalize.

\section{Conclusion}\label{conc}

This work introduced a psychologically-aware, multimodal conversational agent that integrates large language models, knowledge graph reasoning, and prosodic speech analysis to monitor and support students' cognitive engagement and emotional well-being in real time. By bridging semantic understanding and affective inference within a unified architecture, the system transcends the traditional dichotomy between academic instruction and emotional support.

The results of our 8-week pilot study involving 45 students highlight the feasibility of this approach. The proposed model outperformed text-only and prosody-only baselines across all predictive metrics, and a dedicated ablation study showed that both the KG-BERT component and, to a lesser extent given the 30\% voice-interaction rate, the prosodic component contribute measurably to this gain. The model also demonstrated strong alignment with validated psychometric indicators, with statistically significant, large-effect-size reductions in stress and anxiety and notable gains in academic motivation, corrected for multiple comparisons. These findings support the premise that emotionally intelligent educational agents can play a meaningful role in improving learners' psychological resilience and academic engagement, while underscoring that this evidence is exploratory and pilot-scale rather than conclusive.

Beyond technical performance, this research underscores a broader paradigm shift-from automation-centric tutoring systems toward adaptive, ethically grounded, and student-sensitive intelligence. By embedding emotional awareness into real-time pedagogical interventions, our system represents a step toward more inclusive, human-centered learning technologies, contingent on further validation at larger scale and with controlled study designs.

Future work will extend this architecture to longer-term deployments and substantially larger and more diverse learner populations, explicitly addressing the statistical power and generalizability limitations identified in this pilot. We also plan to integrate additional modalities (e.g., gaze tracking, facial expression analysis), increase the share of voice-enabled interactions to strengthen the multimodal evaluation, and investigate privacy-preserving strategies for multimodal data processing, including federated learning and on-device inference. Such directions are crucial to ensuring that emotionally responsive agents remain transparent, scalable, rigorously evaluated, and socially responsible.

In sum, this study contributes to the emerging generation of educational AI systems that not only inform, but also empathize, while explicitly reporting the ablation results, effect sizes, and reliability statistics needed to assess the robustness of this contribution-laying the groundwork for intelligent tutors capable of understanding learners not just as students, but as whole individuals.

\backmatter

\bibliography{sn-bibliography}

\end{document}